\documentclass{ieeeaccess}
\usepackage{cite}
\usepackage{amsmath,amssymb,amsfonts}
\usepackage{algorithmic}
\usepackage{graphicx}
\usepackage{textcomp}
\usepackage{subcaption}
\usepackage{tabularx}
\usepackage{booktabs}       

\usepackage{amsmath}
\usepackage{amssymb}
\usepackage{amsmath,bm}
\usepackage{color}
\usepackage{xcolor,soul}
\usepackage{algorithm}
\usepackage{multirow}

\newcommand{\eg}{e.g.}
\newcommand{\ie}{i.e.}

\newcommand{\etal}{\textit{et al.}}

\setstcolor{red}
\def\revised{\textcolor{black}}

\def\BibTeX{{\rm B\kern-.05em{\sc i\kern-.025em b}\kern-.08em
    T\kern-.1667em\lower.7ex\hbox{E}\kern-.125emX}}
\begin{document}
\history{Date of publication xxxx 00, 0000, date of current version xxxx 00, 0000.}
\doi{10.1109/ACCESS.2017.DOI}

\title{Exploring Global Diversity and Local Context for Video Summarization}
\author{
\uppercase{Yingchao Pan}\authorrefmark{1,2}, 
\uppercase{Ouhan Huang\authorrefmark{1,3}},
\uppercase{Qinghao Ye\authorrefmark{1,4}},
\uppercase{Zhongjin Li\authorrefmark{1}},
\uppercase{Wenjiang Wang\authorrefmark{1}},
\uppercase{Guodun Li\authorrefmark{1}},
\uppercase{And Yuxing Chen\authorrefmark{1}}
}
\address[1]{School of Computer Science and Technology, Hangzhou Dianzi University, Hangzhou 310018, China}
\address[2]{University College London, London, WC1E 6BT, UK}
\address[3]{Key Laboratory for Information Science of Electromagnetic Waves (MoE), Fudan University, Shanghai 200433, China}
\address[4]{Department of Computer Science and Engineering, University of California, San Diego, CA 92037, USA}
\tfootnote{This work was supported by the National Natural Science Foundation of China (No. 61802095).}

\markboth
{Pan \headeretal: Exploring Global Diversity and Local Context for Video Summarization}
{Pan \headeretal: Exploring Global Diversity and Local Context for Video Summarization}

\corresp{Corresponding author: Zhongjin Li (e-mail: lizhongjin@hdu.edu.cn).}

\begin{abstract}
Video summarization aims to automatically generate a diverse and concise summary which is useful in large-scale video processing. Most of the methods tend to adopt self-attention mechanism across video frames, which fails to model the diversity of video frames. To alleviate this problem, we revisit the pairwise similarity measurement in self-attention mechanism and find that the existing inner-product affinity leads to discriminative features rather than diversified features. In light of this phenomenon, we propose global diverse attention which uses the squared Euclidean distance instead to compute the affinities. Moreover, we model the local contextual information by novel local contextual attention to remove the redundancy in the video. By combining these two attention mechanisms, a video \textbf{SUM}marization model with \textbf{D}iversified \textbf{C}ontextual \textbf{A}ttention scheme is developed, namely SUM-DCA. Extensive experiments are conducted on benchmark data sets to verify the effectiveness and the superiority of SUM-DCA in terms of F-score and rank-based evaluation without any bells and whistles.
\end{abstract}

\begin{keywords}
Video summarization, diversified contextual attention, self-attention, similarity function, video representation.
\end{keywords}

\titlepgskip=-15pt

\maketitle

\section{Introduction}
With the rise of video-sharing websites (\eg, YouTube and Facebook), the demand for video analysis surges rapidly. From a content producer's perspective, it is not delightful experience for them to process the long videos. Under this circumstance, the automatic video processing techniques are needed desperately. Video summarization is one of the techniques for handling the massive video data, which removes the redundancy by selecting diverse segments from the video as video summary, and automated methods for generating summary are needed to be investigated.

In the past few decades, various approaches \cite{zhang2016video, vaswani2017attention, zhou2018deep, jung2019discriminative, yuan2019cycle, li2021exploring, zhu2022learning, zhao2022hierarchical} have been proposed to automatically summarize untrimmed videos. Some works \cite{zhang2016video, zhao2018hsa, mahasseni2017unsupervised, zhou2018deep} leverage the Recurrent Neural Networks (RNNs) \cite{giles1994dynamic} and Long Short-Term Memory (LSTM) \cite{hochreiter1997long} for video summarization by modeling the temporal information and show great success. However, these models would fail to handle long videos since recurrent models are not able to model long-range dependency across video frames. This is because the recurrent models tend to suffer serious decay of the history information in terms of long sequences \cite{venugopalan2015sequence}. Recently, attention-based methods \cite{ji2019video, li2021exploring, zhao2022hierarchical, zhu2022learning} have been proposed to alleviate this problem by directly computing the pairwise matrix over the whole video sequence. However, there are several drawbacks: 1) pure self-attention mechanism over all video frames cannot model the diversified feature representation thus is not suitable for video summarization; 2) local temporal cues are unexplored for identifying the most representative essential in local context. 

For the former bottleneck, previous approaches \cite{fajtl2018summarizing, zhao2022hierarchical} tend to adopt self-attention mechanism or multi-head attention mechanism to capture the temporal relation over video frames, which can be implemented by a pairwise frame similarity matrix construction and weighted average summation over all frames. These methods 
simply adopt the dot product as the default pairwise similarity measurement, and we argue that it is not proper for video summarization task. This is because the frame pair with larger magnitude of the weight would suppress the representation of other frames therefore producing the discriminative feature representation for the whole video. But for video summarization, a good summary should reflect diversified semantic information of video, which cannot be satisfied by the dot product similarity measurement. In consequence, we develop the global diverse attention to quantify the importance of each video frame and simultaneously promote the diversity among these frames. In concrete, we find that the choice for pairwise similarity measurement in the pairwise relation modelling is vital. We use $L_2$ similarity to substitute dot product as the similarity measuring function, which leads to more diversified feature representations. Besides, the proposed global diverse attention mechanism shares the similar computation with pure self-attention mechanism by matrix operations, and can be fully optimized by GPU parallel acceleration.

For the latter bottleneck, most existing methods \cite{fajtl2018summarizing, li2021exploring} tend to model the video relation globally while local temporal evolution across consecutive frames is not adequately exploited. Inspired by the concise characteristic of video summarization, the most representative information within a short video segment should be identified and extracted to reduce the redundancy. For further illustration, the beginning of the event and the ending together would foreshadow the happening of the event which should be included in the summary. To this end, we propose a local contextual attention mechanism to identify the discriminative features by modelling the local context information. In particular, the pairwise similarities between an anchored frame and its adjacent frames are computed, then the local contextual feature is generated by weighted aggregation of the adjacent frames. Therefore, the local contextual feature not only includes the representation of the original frame but also integrates local dependency among adjacent frames. In a nutshell, by combining global diverse attention and local contextual attention, we formulate a Diversified Contextual Attention (DCA) scheme and propose a model named SUM-DCA to address the above limitations, which we believe are significant signs of progress for video summarization.

\begin{figure*}
    \centering
    \includegraphics[width=0.95\textwidth]{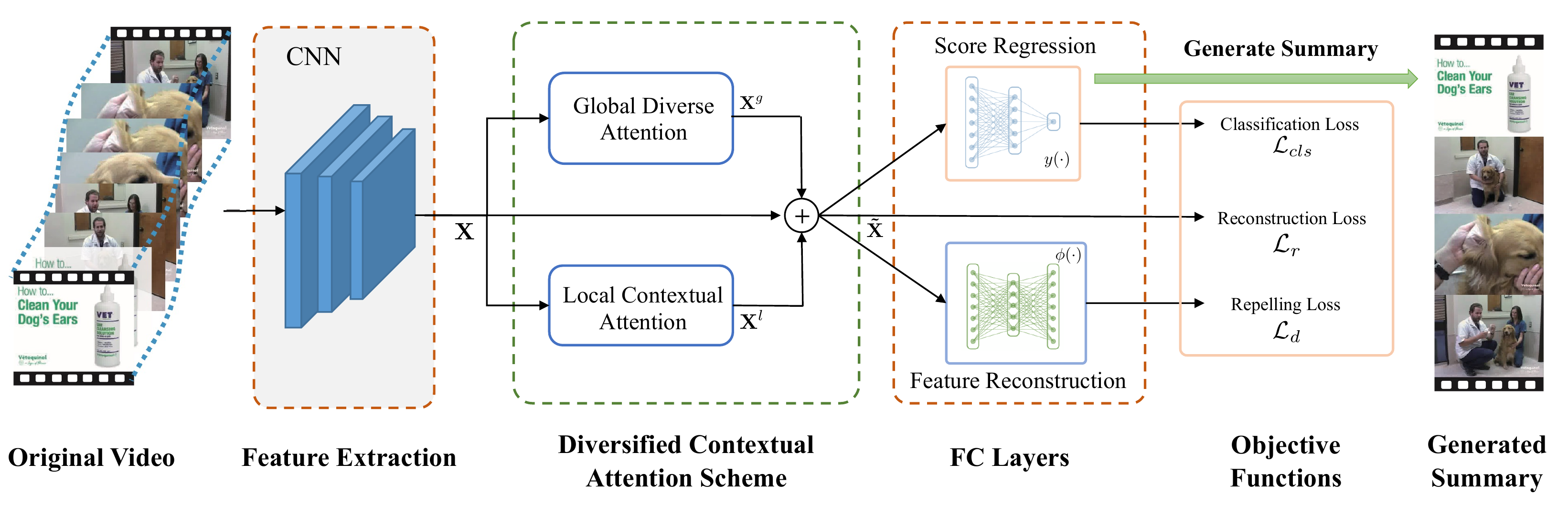}
\caption{Overview of \textbf{SUM-DCA}. Given an input video, SUM-DCA first extracts the features of video frames, then uses diversified contextual attention scheme to model the diversified contextual features $\mathbf{\tilde{X}}$. In the final, it utilizes score regression module to generate a summary automatically. Note that $\mathbf{X}^g$ and $\mathbf{X}^l$ are global diversified features and local contextual features, respectively.}
\label{fig:model}
\end{figure*}

The main contributions can be highlighted as follows:
\begin{itemize}
    \item A diversified contextual attention scheme is developed to model the diversified contextual representation of video by using the pairwise relations among frames, which enables the model to generate diversified and concise summaries.
    \item By delicately selecting the pairwise similarity function that influences the magnitude of frame relations, SUM-DCA is able to generate a diverse representation that conventional self-attention mechanism fails to capture.
    \item Extensive experiments are conducted on the benchmark data sets. The results demonstrate that our model outperforms other competing approaches on SumMe and TVSum datasets.
\end{itemize}

\section{Related Work}
\subsection{Video Summarization}
Video summarization has been widely explored in multimedia analysis with great potential, which can be categorized as two main streams: unsupervised approaches and supervised methods. Our model can be  trained in both supervised and unsupervised fashions.
\subsubsection{Unsupervised Video Summarization}
The unsupervised methods mainly focus on designing heuristic criteria to choose the key shots in terms of representative, diversity, and relevance \cite{zhao2014quasi, ngo2003automatic, liu2002optimization}. VSUMM \cite{de2011vsumm} utilizes k-means to group the visually similar frames into several clusters by color features. Mei \etal \cite{mei20142} treat the summarization problem as $L_{2,0}$-constrained sparse dictionary selection problem and propose the simultaneous orthogonal matching pursuit (SOMP) algorithm. Recently, the deep learning approaches show the great power for unsupervised video summarization. Mahasseni \etal \cite{mahasseni2017unsupervised} put forward an adversarial LSTM networks for generating the summary with summary discriminator. Moreover, Yuan \etal \cite{yuan2019cycle} add cycle-consistency constraint to it to sufficiently align the video and its summary resulting in the comparable summarization performance. Besides, Zhou \etal \cite{zhou2018deep} formulate the summarization problem into a reinforcement learning framework with diversity reward. Rochan \etal \cite{rochan2019video} leverage the pairs of videos to learn the summarization model. Different from these unsupervised methods, we use a simple but intuitive attention model to extract the diversified contextual representations from the video. The pairwise similarities between video frame pairs and the reconstruction process are used to construct objective functions for training. The optimization of our method is more efficient since it does not require adversarial training and reinforcement learning.

\subsubsection{Supervised Video Summarization}
In supervised video summarization, recurrent neural networks (RNN) have been widely adopted in recent years \cite{zhang2016video, zhao2018hsa, li2021exploring}. Zhang \etal \cite{zhang2016video} use bi-directional LSTM to model the temporal dependency of video frames and further introduce determinantal point processes to model the diversity of selected frames. To consider the shot relations within the video, Zhao \etal \cite{zhao2018hsa} develop a hierarchical structure-adaptive rnn to model the intra-shot relation. Instead of using the RNN, SUM-FCN \cite{rochan2018video} uses the 1D fully convolutional neural network to capture the local information of video frames. Besides, Jungji \etal \cite{park2020sumgraph} construct a recurrent graph to model the temporal relation between video frames with residual learning. Jiri \etal \cite{fajtl2018summarizing} introduce the self-attention mechanism for modelling the global information of video frames. In addition, Li \etal \cite{li2021exploring} propose the diverse attention mechanism to capture the global diversity between video frames. Different from previous supervised methods, the proposed model not only captures the local context of the video but also models the global diversity by scrutinizing the self-attention mechanism.

\subsection{Video Highlight Detection}
Video highlight detection \cite{gygli2016video2gif, hong2020mini, ye2021temporal}, a relevant task to video summarization, aims to select the most representative segment from a untrimmed video. Yet the video summarization requires the integrity of the whole video, which does not solely involve the most representative segments. Various studies have been explored in recent years. Gygli \etal \cite{gygli2016video2gif} create the Video-GIF pairs for ranking the video segments to select the highlight segment. To avoid the heavily human annotation, Xiong \etal \cite{xiong2019less} mine the relation between video duration and video highlight by stating the short videos are more likely to contain highlight. To further exploit the video information, Hong \etal \cite{hong2020mini} address the video highlight detection problem not only with visual information but also with audio features. Furthermore, Ye \etal \cite{ye2021temporal} propose a low-rank audio-visual fusion scheme with modelling the temporal dependencies among video segments for better localizing the video highlight segments. In addition, Badamdorj \etal \cite{badamdorj2021joint} introduce noise sentinel to adaptively discount a noisy visual or audio modality during audio-visual fusion.

\section{The Proposed Approach}
In this work, we elaborate the SUMmarization model with Diversified Contextual Attention (SUM-DCA) for video summarization. The overview of SUM-DCA is illustrated in Figure \ref{fig:model}. Specially, the proposed diversified contextual attention scheme contains global diverse attention modelling and local contextual attention modelling, which exploring the diversified frame representation over all frames and mining the local temporal cue in the consecutive frames. Essentially, the global diverse attention models the pairwise relations over the whole video via a pairwise similarity measurement with negative squared Euclidean distance, which performs diversified frame representations with respect to the whole video. Meanwhile, the local contextual attention is able to recognize the most representative frame within a local region by modelling the local temporal contextual information. Finally, we explain the optimization of SUM-DCA and the details of inference.


\subsection{Global Diverse Attention}
\label{sec:gda}
Previous methods \cite{zhang2016video, zhao2018hsa, mahasseni2017unsupervised} estimate the frame importance for video summary directly, without capturing the diversity of the selected frames which is a pivotal characteristic of video summary. In consequence, we propose global diverse attention that exploits the pairwise relations from the video frames for encoding the diversified frame features.

\begin{figure}
    \centering
    \includegraphics[width=0.85\linewidth]{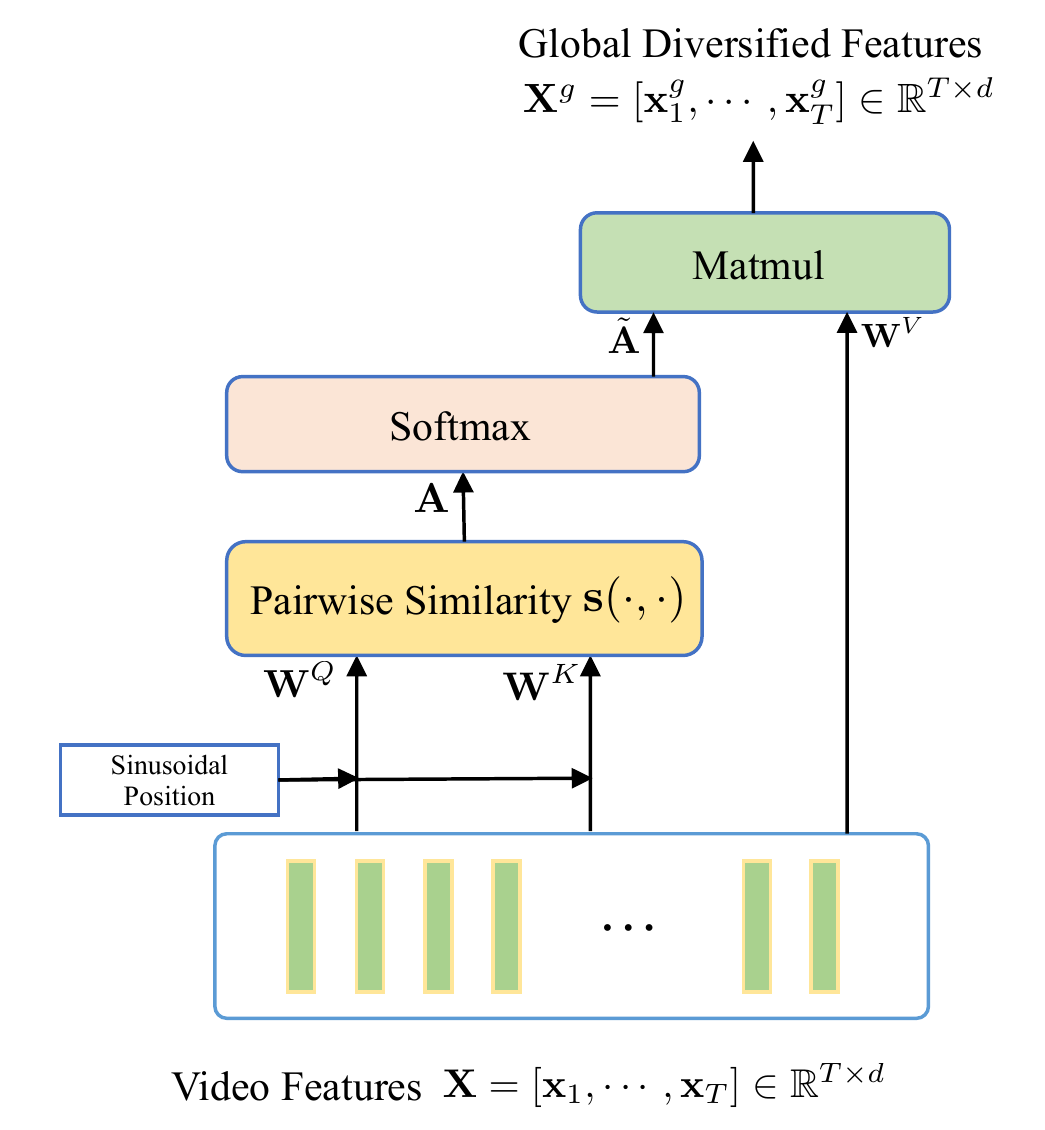}
\caption{The global diverse attention mechanism. $\bf{W}^Q, \bf{W}^K$, and $\bf{W}^V$ are trainable projection matrices that project the video features $\bf{X}$ into different subspaces. The pairwise similarity measurement $\bf{s}(\cdot, \cdot)$ quantifies the similarities between different frame pairs, which acts as the important role in generating different representation.}
\label{fig:gda}
\end{figure}

Given a video $\mathcal{V} = \{\mathbf{v}_1, \cdots, \mathbf{v}_T\}$ with $T$ frames, a pre-trained CNN network, e.g. GoogLeNet \cite{szegedy2015going}, is utilized to extract the corresponding frame features $\bf{X} = [\mathbf{x}_1, \cdots, \mathbf{x}_T] \in \mathbb{R}^{T \times d}$. As depicted in Figure \ref{fig:gda}, the pairwise relation matrix $\bf{A} \in \mathbb{R}^{T\times T}$ that reveals the underlying temporal relations is derived, and each entry $\bf{A}_{ij}$ is the measurement of the similarity between $i$-th frame and $j$-th frame, which is calculated as:
\begin{equation}
    \bf{A}_{ij} = \frac{\bf{s}(\bf{W}^Q\bf{x}_i, \bf{W}^K\bf{x}_j)}{\sqrt{q}},
\label{eq:global_attention}
\end{equation}
where $\bf{s}(\cdot, \cdot):\mathbb{R}^d \times \mathbb{R}^d \rightarrow \mathbb{R}$ is the pairwise similarity measurement. $\bf{W}^Q \in \mathbb{R}^{d\times d}$ and $\bf{W}^K \in \mathbb{R}^{d\times d}$ are learnable linear projection matrices. $q$ is the scaling factor, and we set $q = d$ empirically to avoid the small gradient during the back-propagation. Then, the global diverse attention weights are obtained through applying the softmax normalization as follows:
\begin{equation}
    \tilde{\bf{A}}_{ij} = \frac{\exp(\bf{A}_{ij})}{\sum_{r=1}^T \exp(\bf{A}_{rj})}.
\end{equation}
With the normalized global diverse attention weights $\tilde{\bf{A}}$, the encoded global diversified features $\bf{X}^g = [\mathbf{x}^g_1, \cdots, \mathbf{x}^g_T] \in \mathbb{R}^{T \times d}$ can be computed as a weighted sum of projected video features with efficient matrix multiplication operation:
\begin{equation}
    \bf{X}^g = (\bf{X}\bf{W}^V)\tilde{\bf{A}},
\label{eq:global_sum}
\end{equation}
where $\bf{W}^V \in \mathbb{R}^{d\times d}$ is a trainable linear projection parameter. Moreover, we add the positional information $\bf{P} \in \mathbb{R}^{T\times d}$ to the video sequence $\mathbf{X}$ before applying global diverse attention scheme in order to preserve the temporal order information. \revised{In detail, $\bf{P}$ is added to $\bf{X}$ before calculating the pairwise similarity $\bf{A}$.} Following \cite{vaswani2017attention}, we use the sinusoidal positional embedding which is defined as:
\begin{align}
    \mathbf{P}_{i,2j} &= \sin \left( \frac{i}{10000^{2j/d}} \right), \\
    \mathbf{P}_{i,2j+1} &= \cos \left( \frac{i}{10000^{2j/d}} \right).
\end{align}

The pairwise similarity measurement $\bf{s}(\cdot, \cdot):\mathbb{R}^d \times \mathbb{R}^d \rightarrow \mathbb{R}$ is crucial for generating the diverse and informative features as it supports the feature relation modelling for deriving global diverse attention. One of the common choices is the dot product, i.e. $\bf{s}(\bf{u}, \bf{v}) = \bf{u}^T\bf{v}$, which makes our global diverse attention mechanism is equivalent to the standard self-attention mechanism \cite{vaswani2017attention}. However, if we cast the aggregation procedure in Eq.(\ref{eq:global_sum}), it can be observed that the frame pair with a larger magnitude of the attention weight $\tilde{\bf{A}}_{ij}$ would predominately suppress the representation of other frame pairs. This is practical in some tasks, i.e., video classification, since some pixels that involves objects contain more information than meaningless background pixels. Therefore, the pair with larger magnitude would results in a higher impact on feature representation. But for video summarization, which aims to generate a comprehensive and diverse collection of video segments instead of generating the most representative segment of the video. To handle this problem, we choose $L_2$ similarity for $\bf{s}(\bf{u}, \bf{v})$, which is defined as:
\begin{equation}
    \bf{s}(\bf{u}, \bf{v}) = -\|\bf{u} - \bf{v} \|_2^2,
\label{eq:l2}
\end{equation}
where $\|\cdot\|_2$ is the $L_2$ norm. The naive implementation of $L_2$ similarity involves $O(T^2)$ times computation for loop, which is much slower than the dot product which can be simply implemented within matrix multiplication. To accelerate the computation and fully utilize the parallel characteristic of GPU hardware, we decompose the Eq.(\ref{eq:l2}) as:
\begin{align}
    \bf{s}(\bf{u}, \bf{v}) &= -\|\bf{u} - \bf{v} \|_2^2  \nonumber \\
    &= 2\bf{u}^T\bf{v} - \|\bf{u}\|_2^2 - \|\bf{v}\|_2^2,
\end{align}
which shares a similar computation with dot product and can be easily implemented with matrix operations.

\begin{figure}
\centering
\begin{subfigure}{.5\linewidth}
  \centering
  \includegraphics[width=.8\linewidth]{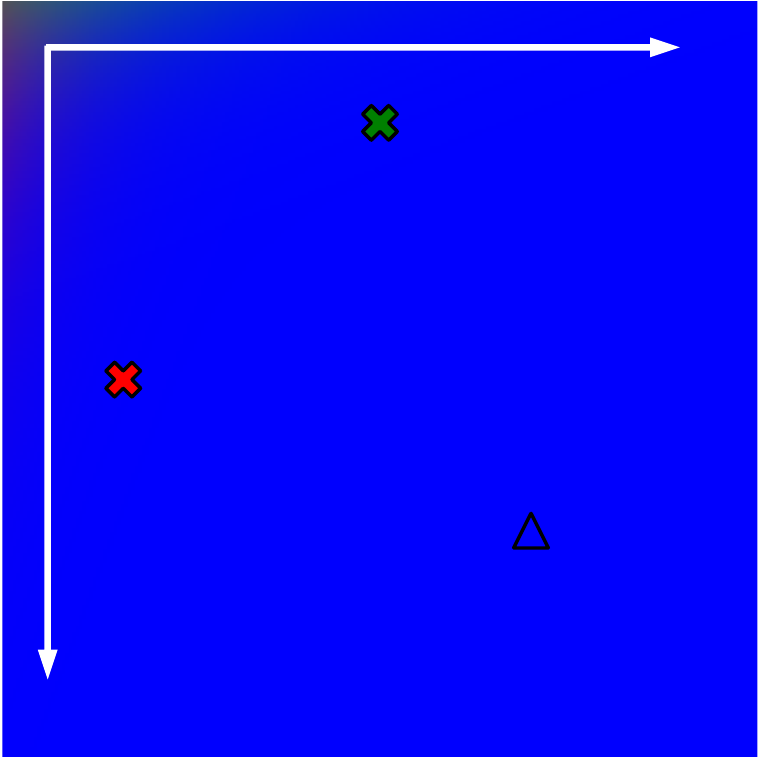}
  \caption{Dot product similarity.}
  \label{fig:dot}
\end{subfigure}%
\begin{subfigure}{.5\linewidth}
  \centering
  \includegraphics[width=.8\linewidth]{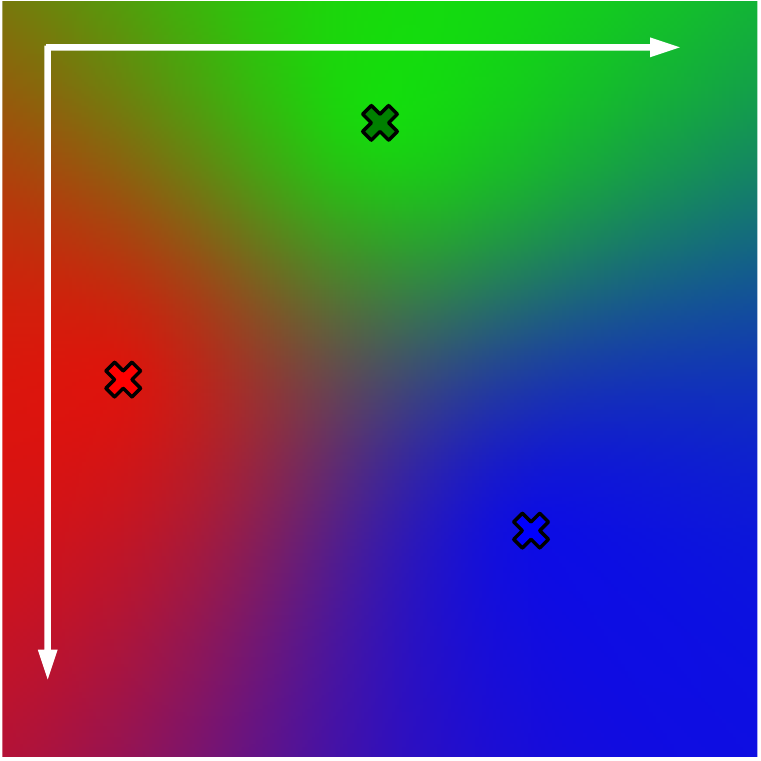}
  \caption{$L_2$ similarity.}
  \label{fig:l2}
\end{subfigure}
\caption{The visualization of softmax contributions (i.e., attention weights $\tilde{\bf{A}}$) from three points. {\color{black}The space is colored as the most similar point under different pairwise similarity measurement.}}
\label{fig:feature_space}
\end{figure}

To give an intuition why $L_2$ similarity can lead to more diversified feature representation than dot product, we visualize a simple but intuitive case under 2-D feature space. As illustrated in Figure \ref{fig:feature_space}, we randomly generate three feature points in the 2D-space, and compute the softmax-contributions (i.e., global diverse attention weights $\tilde{\bf{A}}$) over these three points with respect to any reference point in the space. With the dot product measurement, it can be observed that the blue point dominates the representation of the reference point, which leads to a more discriminative representation. On the other hand, under $L_2$ similarity measurement, each point has the chance to contribute to the feature representation of the reference point resulting in more diversified feature representation over all points.

\subsection{Local Contextual Attention}
Although the global diverse attention models the frame-wise diversity within a video, the temporal cues in the videos are totally ignored. Intuitively, the segment in the summary often contains the most representative essential from the temporal context, i.e., the adjacent video frames around the selected summary frames. Therefore, we argue that temporal contextual information is also pivotal for video summarization. We propose the local contextual attention mechanism to integrate the local information over consecutive frames. The proposed local contextual attention mechanism is able to recognize the most informative frame among the similar adjacent frames as the summary candidates by capturing the temporal contextual cues, thus avoiding the redundancy of the generated summary.

\begin{figure}
    \centering
    \includegraphics[width=0.85\linewidth]{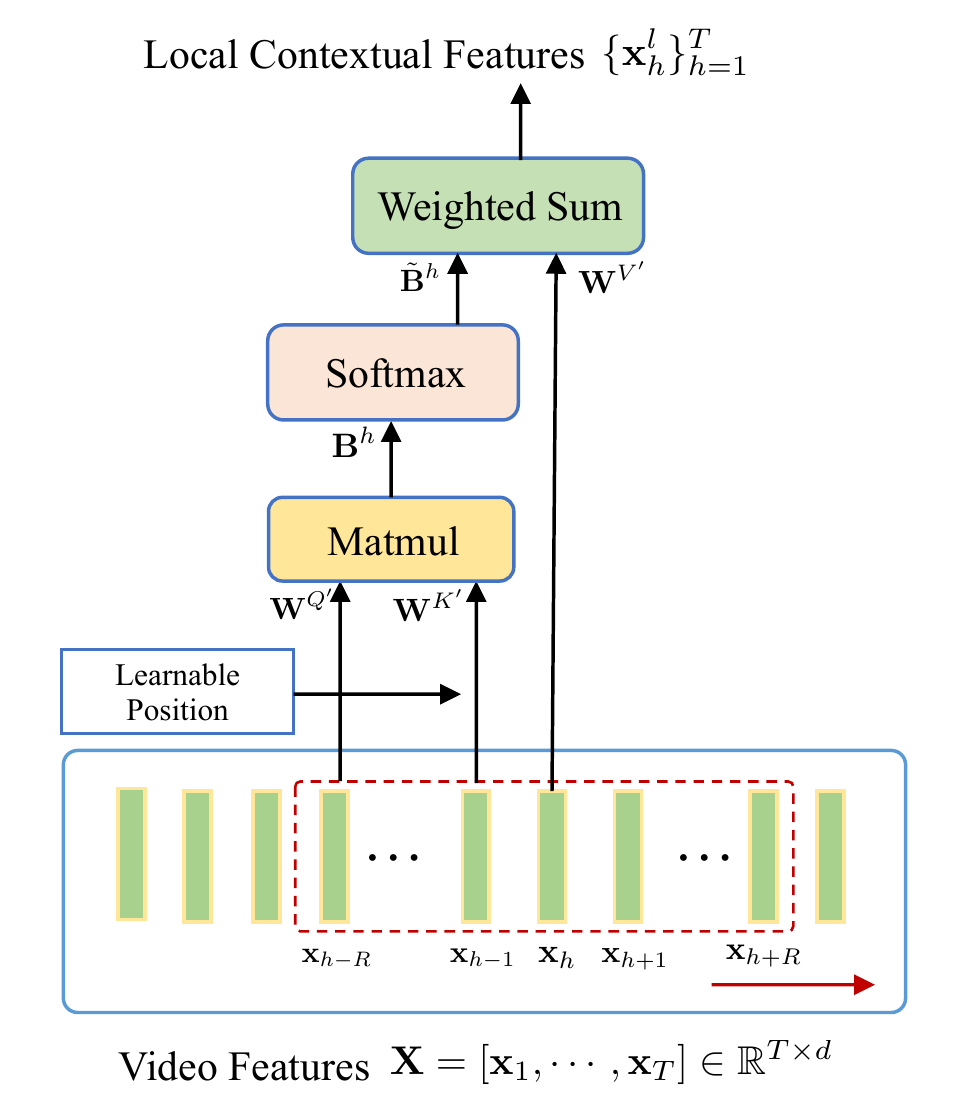}
\caption{The illustration of local contextual attention mechanism. It can aggregate the discriminative frames and mining the temporal cues within the local window.}
\label{fig:lca}
\end{figure}

As shown in Figure \ref{fig:lca}, for each anchored frame $\mathbf{x}_h \in \mathbb{R}^{d}$, where $h \in \{1, 2, \cdots, T\}$, we restrict its attention region to a local scope with its $2R$ adjacent frames:
\begin{equation}
    \hat{\bf{X}}_h = [\mathbf{x}_{h-R}, \cdots, \mathbf{x}_h, \cdots, \mathbf{x}_{h+R}] \in \mathbb{R}^{(2R+1) \times d}.
\end{equation}
Then, the local contextual pairwise matrix $\bf{B}^h \in \mathbb{R}^{(2R+1)\times (2R+1)}$ can be computed through two linear projection, and each entry $\bf{B}^h_{ij}$ can be computed as:
\begin{equation}
    \bf{B}^h_{ij} = \frac{(\bf{W}^{Q'}\bf{x}_{i+h-R})^T (\bf{W}^{K'}\bf{x}_{j+h-R} + \bf{a}_{|i-j|})}{\sqrt{d}},
\label{eq:local_attention}
\end{equation}
where $i, j \in \{0, 1, \cdots, 2R\}$. $\bf{W}^{Q'} \in \mathbb{R}^{d\times d}$ and $\bf{W}^{K'} \in \mathbb{R}^{d\times d}$ are trainable parameters. $[\bf{a}_0, \cdots, \bf{a}_{2R}] \in \mathbb{R}^{(2R+1) \times d}$ are the trainable relative positional embedding vectors that can attend relative distances within the local windows \cite{zhao2022hierarchical}. Next, the local contextual attention weights $\tilde{\bf{B}}^h_{ij}$ within the local window centered at position $h$ is calculated as:
\begin{equation}
    \tilde{\bf{B}}^h_{ij} = \frac{\exp(\bf{B}^h_{ij})}{\sum_{r=0}^{2R} \exp(\bf{B}^h_{rj})}.
\end{equation}

To capture the local contextual information anchored at position $h$ modelled by the local contextual attention weight matrix $\tilde{\bf{B}}^h \in \mathbb{R}^{(2R+1) \times (2R+1)}$, we apply linear projection to the anchored feature $\bf{x}_h$ as:
\begin{equation}
    \bf{x}^l_h = \sum_{r=0}^{2R} \tilde{\bf{B}}^h_{rR} (\bf{W}^{V'} \bf{x}_h),
\label{eq:local_sum}
\end{equation}
where the linear projection matrix $\bf{W}^{V'} \in \mathbb{R}^{d\times d}$ is the parameter to be learned. $\bf{x}^l_h \in \mathbb{R}^d$ is the weighted vector reflecting the local context of the $h$-th frame.

\subsection{The SUM-DCA Model}
By incorporating the global diverse attention mechanism and local contextual mechanism, we can get the global diversified features $\bf{X}^g \in \mathbb{R}^{T \times d}$ and the local contextual features $\bf{X}^l \in \mathbb{R}^{T \times d}$. Then, these two types of features are combined with the original frames representation $\bf{X}$ as:
\begin{equation}
    \tilde{\bf{X}} = \bf{X} + \bf{X}^g + \bf{X}^l,
\label{eq:fuse}
\end{equation}
where $\tilde{\bf{X}} = [\tilde{\mathbf{x}}_1, \tilde{\mathbf{x}}_2, \cdots, \tilde{\mathbf{x}}_T] \in \mathbb{R}^{T\times d}$ is the diversified contextual features for the video $\mathcal{V}$. Then the diversified contextual features are handled by the score regression function $y(\cdot)$ and the embedding function $\phi(\cdot)$. In detail, the score regression function is implemented by two fully-connected layers with the ReLU \cite{nair2010rectified} activation function \revised{and sigmoid function respectively}, which outputs the frame importance score $\bf{y} \in \mathbb{R}^T$.

For training the SUM-DCA model, we employ three loss functions, i.e. classification, repelling, and reconstruction losses. Our model can be trained in both supervised and unsupervised manners. In specific, the supervised setting uses all three losses, and the unsupervised setting only uses the repelling loss and reconstruction loss.

\paragraph{Classification Loss} We use the binary cross-entropy loss for classifying each frame, which is defined as:
\begin{equation}
    \mathcal{L}_{cls} = -\frac{1}{T}\sum_{i=1}^T\left(\hat{y}_i \log y_i + (1-\hat{y}_i)\log (1-y_i)\right),
\end{equation}
where $\hat{y}_i$ is the ground-truth annotation for $i$-th frame, and $y_i$ is $i$-th frame importance score.

\paragraph{Repelling Loss} 
In order to further represent the diversity of the video frames, we employ the repelling loss \cite{mahasseni2017unsupervised} to enhance the diversity of frames which computes the mean value of pairwise cosine similarities between all of $T$ frames as:
\begin{equation}
    \mathcal{L}_d = \frac{1}{T(T-1)}\sum_{i}\sum_{i\ne j}\frac{\phi(\tilde{\mathbf{x}}_i)^T\phi(\tilde{\mathbf{x}}_j)}{\|\phi(\tilde{\mathbf{x}}_i)\|_2 \|\phi(\tilde{\mathbf{x}}_j)\|_2}
\end{equation}
where $\phi(\tilde{\mathbf{x}}_i) \in \mathbb{R}^d$ is the embedding vector of $i$-th frame. 

\paragraph{Reconstruction Loss} A good summary should contain the main content of the video which indicates that the summary has a large reconstruction capacity. Therefore, we use the reconstruction loss to reconstruct the features corresponding to the frames, such as:
\begin{equation}
    \mathcal{L}_r = \frac{1}{T}\sum_{i=1}^T\|\mathbf{x}_i - \varphi(\tilde{\mathbf{x}}_i)\|_2,
\end{equation}
where $\varphi(\cdot)$ is the reconstruction function which is implemented by a two layers fully-connected networks with Sigmoid activation function.

Now, we can obtain the final loss for SUM-DCA in the supervised setting as follows:
\begin{equation}
    \mathcal{L}_{sup} = \mathcal{L}_{cls} + \alpha \mathcal{L}_{d} + \beta \mathcal{L}_{r},
\label{loss:sup}
\end{equation}
where $\alpha$ and $\beta$ are the hyperparameters for controlling the trade-off among three losses. Besides, we also modify the loss to extend SUM-DCA in the unsupervised scenario, i.e. SUM-DCA$_{unsup}$, by omitting the classification loss $\mathcal{L}_{cls}$ as:
\begin{equation}
    \mathcal{L}_{unsup} = \alpha \mathcal{L}_{d} + \beta \mathcal{L}_{r}.
\label{loss:unsup}
\end{equation}
During the training stage, the above loss functions are optimized iteratively, the training procedures are detailed in Algorithm \ref{alg:train}.

\begin{algorithm}[!t]
\begin{algorithmic}[1]
\caption{SUM-DCA Model Training}
\label{alg:train}

\REQUIRE ~~\\
Set of $M$ videos $\{\mathcal{V}_1, \mathcal{V}_2, \cdots, \mathcal{V}_M\}$, learning rate $\eta$.
\ENSURE ~~\\
Learned model parameters: $\Theta$.\\

\STATE Initialize all parameters denoted by $\Theta$ using Xavier.
\STATE Extract frame-level features $\mathbf{X}_m \in \mathbb{R}^{T\times d}$ for all videos.

\REPEAT
	\FOR{$m = 1$ \TO $M$}
    	\STATE Use $\mathbf{X}_m$ to calculate global diversified features $\mathbf{X}^g$ using Eqs.(\ref{eq:global_attention}-\ref{eq:global_sum}).
        \STATE Obtain local contextual features $\mathbf{X}^l$ by \mbox{Eqs.(\ref{eq:local_attention}-\ref{eq:local_sum})}.
        \STATE Get diversified contextual features $\hat{\mathbf{X}}_m$ via Eq.(\ref{eq:fuse}).
		\STATE Calculate frame score $y_i$ by score regression $y(\cdot)$.
		\STATE Obtain transformed feature vector $\phi(\tilde{\mathbf{x}}_i)$ by linear embedding function $\phi(\cdot)$.
		\STATE Compute the loss $\mathcal{L}$ using Eq.(\ref{loss:sup}) or Eq.(\ref{loss:unsup}).
		\STATE $\Theta \leftarrow \Theta - \eta \bigtriangledown_{\Theta} \mathcal{L}(\Theta)$.
	\ENDFOR
\UNTIL{convergence}

\RETURN $\Theta$.

\end{algorithmic}
\end{algorithm}

\subsection{Summary Generation}
For summary generation, a set of key shots is selected by maximizing the frame scores. In specific, we follow \cite{zhang2016video, rochan2019video} to generate a set of changing points using Kernel Temporal Segmentation (KTS) \cite{potapov2014category} therefore dividing the video into total $S$ shots. The summary proportion constraint $l \leq T$ is applied for controlling the length of the generated summary. Then, the key shots are selected by the 0/1 Knapsack algorithm \cite{balas1980algorithm}, which is formulated as:
\begin{equation}
\label{eq:greedy}
\max_{p_i}\,\,  \sum_{i=1}^S p_i s_i, \quad
s.t.~
\begin{cases}
\sum\limits_{i=1}^S p_i l_i \leq l,   \\
s_i = \frac{1}{l_i}\sum\limits_{t=1}^{l_i} y_{i}^t,  \\
p_i \in \{0, 1\},
\end{cases}
\end{equation}
where $s_i$ indicates the average score over the frame within the $i$-th shot generated by KTS, and $l_i$ is the length of $i$-th shot. If $p_i = 1$, the $i$-th shot is chosen to compose the summary. The video summary generation steps are summarized in the Algorithm \ref{alg:test}.

\begin{algorithm}[!t]
\caption{Video Summary Generation}\label{alg:test}
\begin{algorithmic}[1]
\REQUIRE ~~\\
Test video $\mathcal{V}$ and model parameters $\Theta$.\\
	
\ENSURE ~~\\
Video Summary $\mathcal{S}$. \\

\STATE Initialize $\mathcal{S} \leftarrow \emptyset$.
\STATE Extract frame features $\mathbf{X} = [\mathbf{x}_1, \mathbf{x}_2, \cdots, \mathbf{x}_T] \in \mathbb{R}^{T\times d}$ by a pre-trained CNN model.
\STATE Use KTS \cite{potapov2014category} to divide the video $\mathcal{V}$ into $S$ shots $\{\mathbf{S}_i\}_{i=1}^S$.
\STATE Compute global diversified features $\mathbf{X}^g$ using Eqs.(\ref{eq:global_attention}-\ref{eq:global_sum}).
\STATE Calculate local contextual features $\mathbf{X}^l$ by \mbox{Eqs.(\ref{eq:local_attention}-\ref{eq:local_sum})}.
\STATE Get diversified contextual features $\hat{\mathbf{X}}$ via Eq.(\ref{eq:fuse}).
\STATE obtain frame scores $y_i$ for each diversified contextual features $\hat{\mathbf{X}}_i$ with frame score regression module $y(\cdot)$.
\STATE Solve optimal $p_i$ for each shot $\mathbf{S}_i$ through Eq.(\ref{eq:greedy}).
\FORALL{\text{shot $\mathbf{S}_i$ in video $\mathcal{V}$}}
	\STATE \textbf{if} $p_i = 1$, \textbf{then} \\
	\STATE \quad \quad$\mathcal{S} \leftarrow$ $\mathcal{S} \cup \{\mathbf{S}_i\}$.
\ENDFOR
\RETURN $\mathcal{S}$.
\end{algorithmic}
\end{algorithm}

\section{Experiments}

\begin{table}[]
\caption{Three different evaluation settings for TVSum dataset. To evaluate on the SumMe dataset, the position of SumMe and TVSum should be switched.}
\begin{center}
\begin{tabular}{lcc}
\toprule
\textbf{Evaluation Setting}    & \textbf{Training}                           & \textbf{Testing}    \\ \midrule
Canonical (C)  & 80\% TVSum                         & 20\% TVSum \\ \hline
Augmented (A)  & \begin{tabular}[c]{@{}c@{}}80\% TVSum + SumMe \\ + OVP + YouTube\end{tabular} & 20\% TVSum \\ \hline
Transfer (T)   & SumMe + OVP + YouTube              & TVSum      \\
\bottomrule
\end{tabular}
\end{center}
\label{tb:settings}
\end{table}

\begin{table*}[!t]
\begin{center}
\caption{Performance comparison (F-score \%) with supervised methods on SumMe and TVSum. }
\label{tab:CAT}
\begin{tabular}{lcccccc}\toprule
\multirow{2}{*}{Method} & \multicolumn{3}{c}{SumMe}  & \multicolumn{3}{c}{TVSum}     \\ \cmidrule(l{2pt}r{2pt}){2-4} \cmidrule(l{2pt}r{2pt}){5-7}
                        & Canonical & Augment & Transfer & Canonical & Augment & Transfer \\ \midrule
Bi-LSTM~\cite{zhang2016video}                      & 37.6 & 41.6  & 40.7 & 54.2 & 57.9                      & 56.9                 \\
DPP-LSTM~\cite{zhang2016video}                     & 38.6 & 42.9  & 41.8 & 54.7 & 59.6                      & 58.7                 \\
SUM-GAN$_{sup}$~\cite{mahasseni2017unsupervised}   & 41.7 & 43.6  & -    & 56.3 & 61.2                      & -                    \\
DR-DSN$_{sup}$~\cite{zhou2018deep}    & 42.1 & 43.9  & 42.6 & 58.1 & 59.8                      & 58.9                 \\
SUM-FCN~\cite{rochan2018video}                      & 47.5 & 51.1  & 44.1 & 56.8 & 59.2                      & 58.2                 \\
HSA-RNN~\cite{zhao2018hsa}                      & - & 44.1  & - & - & 59.8                      & -                 \\
CSNet$_{sup}$~\cite{jung2019discriminative}     & 48.6 & 48.7  & 44.1 & 58.5 & 57.1                      & 57.4                 \\
VASNet~\cite{fajtl2018summarizing}       & 49.7 & 51.1  & -    & \textbf{61.4} & \textbf{62.4}         & -                \\
HMT \cite{zhao2022hierarchical} & 44.1 & 44.8  & - & 60.1 &  60.3                    &  -                    \\
M-AVS~\cite{ji2019video}                 & 44.4 & 46.1  & -    & 61.0 & \underline{61.8}                      & -                 \\ 
SUM-GDA \cite{li2021exploring}          & \underline{52.8} & \underline{54.4}  & \underline{46.9}   & 58.9 & 60.1         & \underline{59.0}                 \\ 
MHANet \cite{zhu2022learning}          & 51.1 & 52.1  & 45.4   & 61.0 & 61.5         & 55.1                 \\  \hline
SUM-DCA                      & \textbf{54.7} & \textbf{57.7}  & \textbf{49.7} & \underline{61.3} & 61.6         & \textbf{59.8}         \\
\bottomrule
\end{tabular}
\end{center}
\end{table*}

\begin{table*}[!t]
\begin{center}
\caption{Performance comparison (F-score \%) with unsupervised methods on SumMe and TVSum.}
\label{tab:CATunsup}
\begin{tabular}{lcccccc}\toprule
\multirow{2}{*}{Method} & \multicolumn{3}{c}{SumMe}  & \multicolumn{3}{c}{TVSum}     \\ \cmidrule(l{2pt}r{2pt}){2-4} \cmidrule(l{2pt}r{2pt}){5-7}
                        & Canonical & Augment & Transfer & Canonical & Augment & Transfer \\ \midrule
SUM-GAN$_{rep}$~\cite{mahasseni2017unsupervised}   & 38.5 & 42.5  & -    & 51.9 & 59.3                      & -                    \\
SUM-GAN$_{dpp}$~\cite{mahasseni2017unsupervised}   & 39.1 & 43.4  & -    & 51.7 & 59.5                      & -                    \\
DR-DSN~\cite{zhou2018deep}                       & 41.4 & 42.8  & 42.4 & 57.6 &	58.4 					  &	57.8                 \\
CSNet~\cite{jung2019discriminative}   & \textbf{51.3} & \textbf{52.1}  & 45.1 & 58.8 & 59.0            & \textbf{59.2}                 \\
Cycle-SUM~\cite{yuan2019cycle}     & 41.9 & -  & -     & 57.6 & -                      & -                 \\
UnpairedVSN~\cite{rochan2019video}     & 47.5 & -  & -     & 55.6 & -                      & -                 \\
$\text{SUM-GDA}_{unsup}$ \cite{li2021exploring} & 50.0 & 50.2  & \underline{46.3} & \underline{59.6} &     60.5         &  58.8             \\
SumGraph \cite{park2020sumgraph} & 49.8 & \underline{52.1}  & \textbf{47.0}& 59.3 &     \textbf{61.2}                     &  57.6                     \\ \hline
$\text{SUM-DCA}_{unsup}$ & \underline{50.4} & 50.7  & 46.1 & \textbf{60.4} &     \underline{60.5}                     &  \underline{59.1}                   \\
\bottomrule
\end{tabular}
\end{center}
\end{table*}

\subsection{Datasets}
We employed four data sets for this paper, including SumMe \cite{gygli2014creating}, TVSum \cite{song2015tvsum}, Open Video Project (OVP) \cite{de2011vsumm}, and YouTube \cite{de2011vsumm}. SumMe consists of 50 videos within 1-5 minutes length that have various topics, such as news, documentary, how to videos, etc. SumMe data set is a collection of 25 user videos that record different events including holidays, history, and sports. The length of videos in SumMe varies from 1.5 minutes to 6.5 minutes. For YouTube and OVP data sets, 39 and 50 videos are collected with cartoons, news, and sports topic, respectively. These data sets are diverse in terms of the content and come with different type of annotations, \ie, shot-level scores for SumMe and frame-level scores for TVSum. We use SumMe and TVSum data sets for training and evaluation. The YouTube and OVP data sets are only used during training.

\subsection{Evaluation Metrics}
Following \cite{zhang2016video, zhou2018deep, mahasseni2017unsupervised}, we use the $F$-measure score to evaluate the similarities of the generated summary among user annotated summary. Formally, the precision of generated summary is computed as $\mathcal{P} = \frac{|\mathcal{S}_{user}\bigcap \mathcal{S}_{machine}|}{|\mathcal{S}_{machine}|}$, and the recall is calculated as $\mathcal{R} = \frac{|\mathcal{S}_{user}\bigcap \mathcal{S}_{machine}|}{|\mathcal{S}_{user}|}$, where $\mathcal{S}_{user}$ and $\mathcal{S}_{machine}$ are the user annotated summary and the generated summary respectively. Then the $F$-score is defined by:
\begin{equation}
    F = \frac{2\times \mathcal{P} \times \mathcal{R}}{\mathcal{P} + \mathcal{R}} \times 100 \%.
\end{equation}

Although $F$-score is widely used in the evaluation among various video summarization methods, Mayu \etal \cite{otani2019rethinking} stated that the random method can achieve similar or comparable results due to the well-designed post-processing. To alleviate this problem, the rank correlation statistics, i.e. Kendall’s coefficient $\tau$ and Spearman’s coefficient $\rho$ are utilized to measure the similarity between ranked human annotated scores and prediction scores for the video frames. Therefore, we also evaluate the performance among different methods with these metrics.

\subsection{Evaluation Settings}
For SumMe and TVSum, following the evaluation settings in \cite{zhang2016video, yuan2019cycle, li2021exploring}, we adopt the Canonical (C), Augmented (A), and Transfer (T) settings as described in Table \ref{tb:settings}. The dataset is randomly split into two disjoint sets: 80\% for training and 20\% for testing. Five-folds cross validation is adopted for avoiding the randomness, and we report the average result among five testing splits.

\subsection{Implementation Details}
For fair comparison, each video is sub-sampled at 2 frames per second rate in order to remove the redundancy. Then, the features $\mathbf{x}_i \in \mathbb{R}^{1024}$ of sampled frames $\mathbf{v}_i$ are extracted from the output of pool-5 layer of GoogLeNet \cite{szegedy2015going} pre-trained on ImageNet \cite{russakovsky2015imagenet}. \revised{Following Zhang \etal\cite{zhang2016video}, we transform the frame-level annotated importance scores with 0/1 Knapsack algorithm into the binary ground-truth annotations since the binary ground-truth would indicate whether the frame belongs to summary or not.} The number of hidden units for each linear projection layer is set to 1024. Our model is trained with Adam optimizer \cite{kingma2014adam} with the $L_2$ weight decay coefficient $10^{-5}$. We empirically choose the hyperparameter $\alpha = 0.1$ and $\beta = 1$ \revised{for both Eq.(\ref{loss:sup}) and (\ref{loss:unsup})}. The initial learning rate is set to $1\times 10^{-4}$ for SumMe, and $1\times 10^{-4}$ for TVSum dataset. For each dataset and each setting, we train our method for 200 epochs on a machine with a NVIDIA Titan Xp GPU using PyTorch platform \cite{NEURIPS2019_9015}.

\subsection{Quantitative Results}
We compare the proposed SUM-DCA with several state-of-the-art methods including both supervised and unsupervised approaches in terms of three settings described in Table \ref{tb:settings}.

For SumMe and TVSum, we compare our SUM-DCA with supervised approaches including Bi-LSTM \cite{zhang2016video}, DPP-LSTM \cite{zhang2016video}, SUM-GAN$_{sup}$ \cite{mahasseni2017unsupervised}, DR-DSN$_{sup}$ \cite{zhou2018deep}, SUM-FCN \cite{rochan2018video}, HSA-RNN \cite{zhao2018hsa}, VASNet \cite{fajtl2018summarizing}, CSNet$_{sup}$ \cite{jung2019discriminative}, HMT \cite{zhao2022hierarchical}, M-AVS \cite{ji2019video}, SUM-GDA \cite{li2021exploring}, and MHANet \cite{zhu2022learning}, while the unsupervised methods include SUM-GAN, DR-DSN, CSNet, Cycle-SUM \cite{yuan2019cycle}, UnpairedVSN \cite{rochan2019video}, and SumGraph \cite{park2020sumgraph}. Especially, these compared methods can be categorized in three aspects, \ie, RNN-based methods, GAN-based methods, and attention-based methods.

\begin{table}
\begin{center}
\caption{Performance comparison of rank statistics $\tau$ and $\rho$ among different approaches. This experiment uses TVSum data set under canonical setting.}
\label{tab:rank}
\begin{tabular}{lcc}
\toprule
Methods     & Kendall’s $\tau$  & Spearman’s $\rho$ \\ \midrule

DPP-LSTM \cite{zhang2016video} & 0.042 & 0.055 \\
DR-DSN \cite{zhou2018deep} & 0.020 & 0.026 \\
HSA-RNN \cite{zhao2018hsa} & 0.082 & 0.088 \\
VASNet \cite{vaswani2017attention} & 0.082 & 0.088 \\
SUM-GAN~\cite{mahasseni2017unsupervised} & -0.054 & -0.070 \\
SUM-FCN \cite{rochan2018video} & 0.011 & 0.014 \\
CSNet \cite{jung2019discriminative} & 0.070 & 0.091 \\
HMT \cite{zhao2022hierarchical} & 0.096 & 0.107 \\
SUM-DCA$_{unsup}$ & \underline{0.106}  & \underline{0.126} \\
SUM-DCA & \textbf{0.124}  & \textbf{0.152}  \\ \hline \hline
Random     & 0.000  &  0.000  \\ 
Human   & 0.177  &  0.204 \\ 
\bottomrule
\end{tabular}
\end{center}
\end{table}

Table \ref{tab:CAT} summarizes the experimental results of different supervised approaches. We can find that our model achieves the best performance on the SumMe data set in all three settings, while obtaining the best generalization performance in terms of Transfer setting on both data sets. In particular, our method yields a better performance at least by 6.1\% higher than the RNN-based method (\ie, Bi-LSTM \cite{zhang2016video}, DPP-LSTM \cite{zhang2016video}, DR-DSN$_{sup}$ \cite{zhou2018deep}, HSA-RNN \cite{zhao2018hsa}, and CSNet$_{unsup}$ \cite{jung2019discriminative}) on SumMe under canonical setting and 9.0\% under the augmented setting. This is because the RNN-based methods fail to model the long-term dependency, thus cannot capture the long-term context information for effectively summarizing the video. Besides, we can observe that SUM-DCA outperforms the attention-based methods (\ie, M-AVS \cite{ji2019video}, VASNet \cite{fajtl2018summarizing}, HMT \cite{zhao2022hierarchical}, SUM-GDA \cite{li2021exploring}, and HMANet \cite{zhu2022learning}) by a large margin due to the proper choice of the pairwise similarity measurement $\mathbf{s}(\cdot, \cdot)$. We demonstrate attention mechanism is crucial for modeling the global information of the video in terms of summarization. For example, M-AVS \cite{ji2019video} utilizes the encoder-decoder structure to model the additive attention for measuring the similarity, and VASNet \cite{fajtl2018summarizing} adopts the pure self-attention mechanism for encoding the global similarity information. These approaches, however, solely model the similarity among video frames resulting in the discriminative video features. SUM-GDA \cite{li2021exploring} models the global dissimilarities of the video frames instead of the similarity, and achieves a relative gain of 3.1 \% on SumMe with the canonical setting. Our proposed SUM-DCA not only yields the diversified frame features but also captures the local context among several frames, therefore leading to the higher performance than SUM-GDA in all aspects.

Table \ref{tab:CATunsup} presents the the experimental results of unsupervised methods. It can be observed that our method is able to achieve the comparable performance among competing unsupervised approaches. Typically, compared to the GAN-based methods (\ie, SUM-GAN \cite{mahasseni2017unsupervised}, Cycle-SUM \cite{yuan2019cycle}, and UnpairedVSN \cite{rochan2019video}), our pure attention model achieves a relative gain of 2.9\% on SumMe data set and 1.6\% on TVSum with the canonical setting. In addition, we also notice that the difference of performance over three settings on TVSum is relatively smaller among all of the methods than on SumMe. This might be due to the fact that SumMe is more challenging and adopts the highest F-score among several users which shows more targeted when doing evaluation, while TVSum adopts the average F-scores among several users, and the users are not likely to make the consistent agreement.

Moreover, we evaluate the summarization performance by the rank-based evaluation, which compute the correlation between the predict probabilities and the annotated importance scores by human. Two rank-based metrics are employed in this paper, \ie Kendall's $\tau$ and Spearman's $\rho$. The results are summarized in the Table \ref{tab:rank}. As we can observe in the table, the performance of random selection and human annotation are the lowest and the highest respectively. In particular, our SUM-DCA surpasses other state-of-the-art method by a significant margin. Besides, with the help of annotation, the SUM-DCA performs better than SUM-DCA$_{unsup}$ in terms of both $\tau$ and $\rho$. Overall, the results in Table \ref{tab:rank} indicate the advantages of the proposed SUM-DCA with the following aspects: 1) The proposed global diverse attention mechanism can capture the global dependencies among frames while model the diversified frame representation. 2) The local contextual attention is able to integrate the local information over consecutive frames, which is useful for avoiding the duplication during summary generation.

\subsection{Ablation Studies}

\begin{table}
\begin{center}
\caption{Ablation on Global Diverse Attention (GDA) and Local Contextual Attention (LCA) mechanisms in our SUM-DCA model. This experiment uses SumMe and TVSum data sets under canonical setting.}
\label{tab:gda_lca}
\begin{tabular}{ccccc}
\toprule
Exp No. & GDA & LCA & SumMe  & TVSum \\ \midrule
1   &   &   & 44.5  & 55.7 \\
2   & \checkmark   &   & 52.8  & 60.3 \\
3   &   & \checkmark   & 51.3  & 60.7 \\
4   & \checkmark   & \checkmark   & \textbf{54.7}  & \textbf{61.3} \\
\bottomrule
\end{tabular}
\end{center}
\end{table}

\begin{table}
\begin{center}
\caption{Ablation on the pairwise similarity measurement $\mathbf{s}(\cdot, \cdot)$ in global diverse attention mechanism. This experiment is conducted on SumMe and TVSum data sets under canonical setting.}
\label{tab:similarity}
\begin{tabular}{lcc}
\toprule
Similarity Function $\mathbf{s}(\cdot, \cdot)$ & SumMe  & TVSum \\ \midrule
Dot Product  & 51.2  & 60.1 \\
Cosine Similarity  & 51.9  & 60.7 \\
$L_2$ Similarity  & \textbf{54.7}  & \textbf{61.3} \\
\bottomrule
\end{tabular}
\end{center}
\end{table}

\begin{table*}[!t]
\begin{center}
\caption{Variations in performance (F-score \%) by changing the neighbor size $R$ on SumMe and TVSum.}
\label{tab:local}
\begin{tabular}{ccccccc}\toprule
\multirow{2}{*}{Neighbor Size $R$} & \multicolumn{3}{c}{SumMe}  & \multicolumn{3}{c}{TVSum}     \\ \cmidrule(l{2pt}r{2pt}){2-4} \cmidrule(l{2pt}r{2pt}){5-7}
                        & Canonical & Augment & Transfer & Canonical & Augment & Transfer \\ \midrule
1	& \textbf{54.7}	& 52.0	& 48.6	& 60.9	& 61.5	& 59.6     \\
2	& 51.1	& 51.8	& \textbf{49.7}	& 60.9	& \textbf{61.6}	& \textbf{59.8}     \\
3	& 53.2	& 56.1	& 49.5	& 60.8	& 61.5  & 59.8     \\
4	& 54.0	& \textbf{57.7}	& 49.5	& 61.1	& 61.4	& 59.8     \\
5	& 52.5	& 55.5	& 48.4	& 60.8	& 61.4	& 59.4     \\
6	& 52.0	& 56.7	& 49.3	& 61.1	& 61.3	& 59.7     \\
7	& 54.2	& 55.4	& 49.6	& 61.1	& 61.2	& 59.1     \\
8	& 53.7	& 52.0	& 49.3	& \textbf{61.3}	& 61.4	& 59.6     \\
\bottomrule
\end{tabular}
\end{center}
\end{table*}

\begin{table*}[!t]
\begin{center}
\caption{Variations in performance (F-score \%) by training SUM-DCA with difference losses on SumMe and TVSum.}
\label{tab:loss}
\begin{tabular}{cccccccccc}\toprule
\multirow{2}{*}{Exp No.} & \multirow{2}{*}{$\mathcal{L}_{cls}$} & \multirow{2}{*}{$\mathcal{L}_{d}$} & \multirow{2}{*}{$\mathcal{L}_{r}$} & \multicolumn{3}{c}{SumMe}  & \multicolumn{3}{c}{TVSum}     \\ \cmidrule(l{2pt}r{2pt}){5-7} \cmidrule(l{2pt}r{2pt}){8-10}
      &   &   &  & Canonical & Augment & Transfer & Canonical & Augment & Transfer \\ \midrule
1   & \checkmark  &   &	  &  52.1 &	54.3 &	47.1 &	57.6 &	58.8 &	55.7    \\
2   &   & \checkmark  &	  &  45.4 &	49.1 &	43.2 &	57.5 &	58.8 &	56.6    \\
3   &   &   & \checkmark  &  46.3 &	49.8 &	44.5 &	58.1 &	59.7 &	58.8   \\ \hline
4   & \checkmark  & \checkmark  &	  & 53.7 &	55.4 &	48.5 &	61.2 &	61.1 &	59.7     \\
5   & \checkmark  &   &	\checkmark  & 54.7 &	54.9 &	49.3 &	61.0 &	61.5 &	59.7     \\
6   &   & \checkmark  &	\checkmark  & 50.4 &	50.7 &	46.1 &	60.4 &	60.5 &	59.1     \\ \hline
7   & \checkmark  & \checkmark  & \checkmark  & \textbf{54.7} &	\textbf{57.7} &	\textbf{49.7} &	\textbf{61.3} &	\textbf{61.6} &	\textbf{59.8}     \\
\bottomrule
\end{tabular}
\end{center}
\end{table*}

In this section, we perform ablation study for verifying the effectiveness of each component in our model under different conditions.

\subsubsection{Impact of Global Contextual Attention Scheme}
Firstly, to examine the influences of global contextual attention scheme, we conducted ablation on our proposed global contextual attention scheme and Table \ref{tab:gda_lca} tabulates the main results. It suggests that there is a significant performance drop when the global diverse attention is not used. This indicates that the diversified frame representation is essential for generating a satisfied summary. Besides, the local contextual attention alone (Exp No.3) also vital for summarization since it utilizes the local temporal cues thus ruling out the redundancy among adjacent frames. Moreover, by applying global contextual attention (Exp No.4), our method achieves the highest performance in both two data sets.

\subsubsection{Impact of Pairwise Similarity Measurement $\mathbf{s}(\cdot, \cdot)$}
Secondly, we have proved that $L_2$ similarity is more suitable for $\mathbf{s}(\cdot, \cdot)$. The results in Table \ref{tab:similarity} demonstrate that $L_2$ similarity improves the performance with negligible overhead on both SumMe and TVSum. In particular, we can find that the cosine similarity performs better than the plain dot product since the normalization of vectors suppresses the model to generate the discriminative representation to some extent. In addition, the $L_2$ similarity outperforms the rest of the two similarity measurements by at least 2.8\% on SumMe and 0.6\% on TVSum. This is consistent with the content of Section \ref{sec:gda} - the $L_2$ similarity can lead to more diversified feature representation thus boosting the summarization performance.

\subsubsection{Impact of Neighbor Size $R$}
Then, we investigate the effect of varying the neighbor size $R$ for local contextual attention, as described in Table \ref{tab:local}. It can be observed that the small value of $R$ (\ie, $R = 2$ or $R = 3$) would gain the best performance. This suggests local contextual attention within small region would find the most discriminative features among the similar adjacent frames, which fully utilizes the local temporal cue and avoids the redundancy during the summary generation. Besides, for the large $R$, it might model excessive context which hinders the model from learning discriminative frame features among similar frames, thus leading to degenerated summary performance.

\subsubsection{Impact of Loss Terms}
Furthermore, we ablate the contributions of individual loss terms and the performances are shown in Table \ref{tab:loss}. As can be seen from the table, for the supervised setting, the classification loss $\mathcal{L}_{cls}$ alone (Exp No.1) yields the lowest F-scores. Adding the repelling loss (Exp No.4) and reconstruction loss (Exp No.5) would improve performance significantly, while combining all three losses (Exp No.7) leads to the best performance and surpass the model with $\mathcal{L}_{cls}$ only by at least 2.6\% on SumMe and 3.7\% on TVSum respectively. However, in the unsupervised setting, as the ground truth annotations cannot be used, the classification loss is removed. This causes a significant degrade in the performance compared with models in a supervised setting.

\begin{figure*}
\centering
\begin{subfigure}{.8\linewidth}
  \centering
  \includegraphics[width=\linewidth]{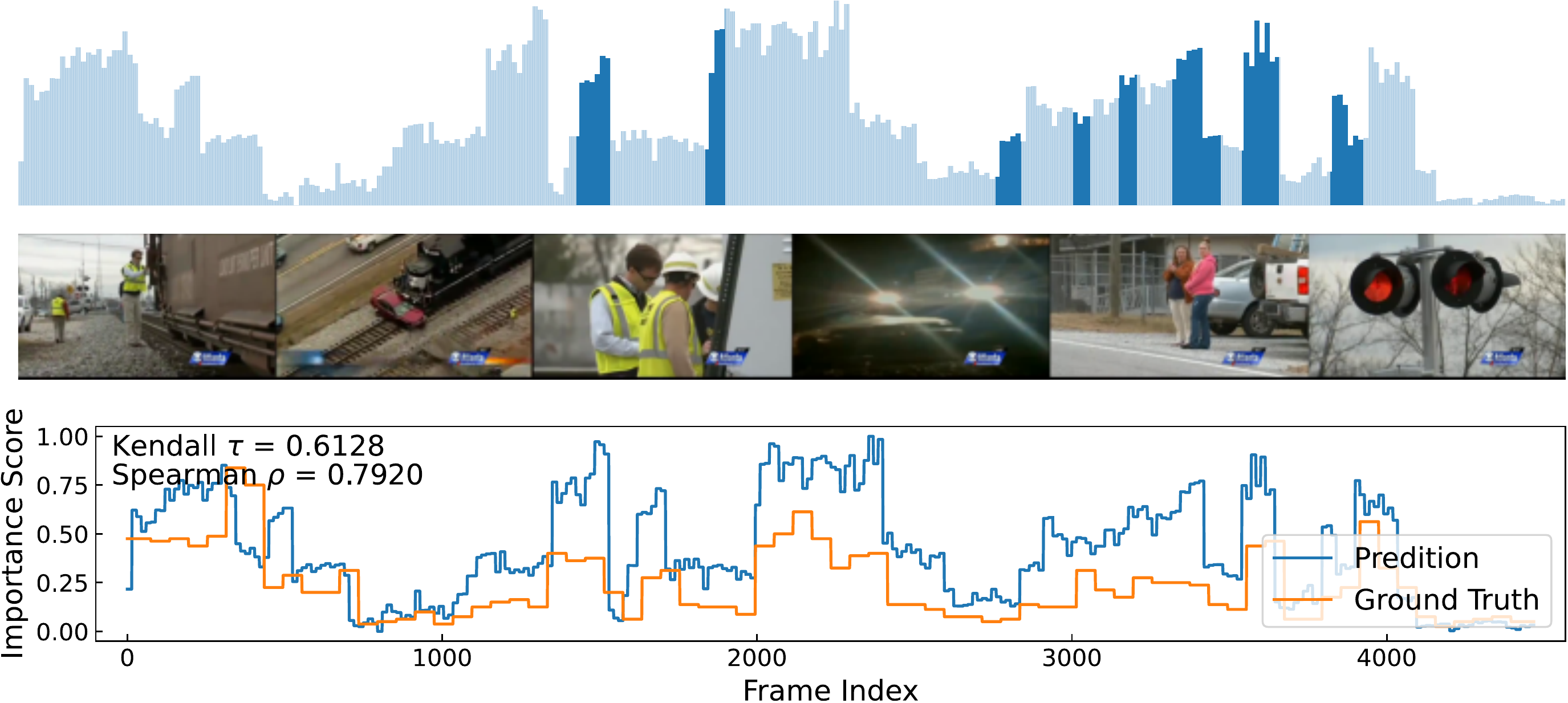}
  \caption{Video name: sTEELN-vY30.mp4}
  \label{fig:viz1}
\end{subfigure}

\begin{subfigure}{.8\linewidth}
  \centering
  \includegraphics[width=\linewidth]{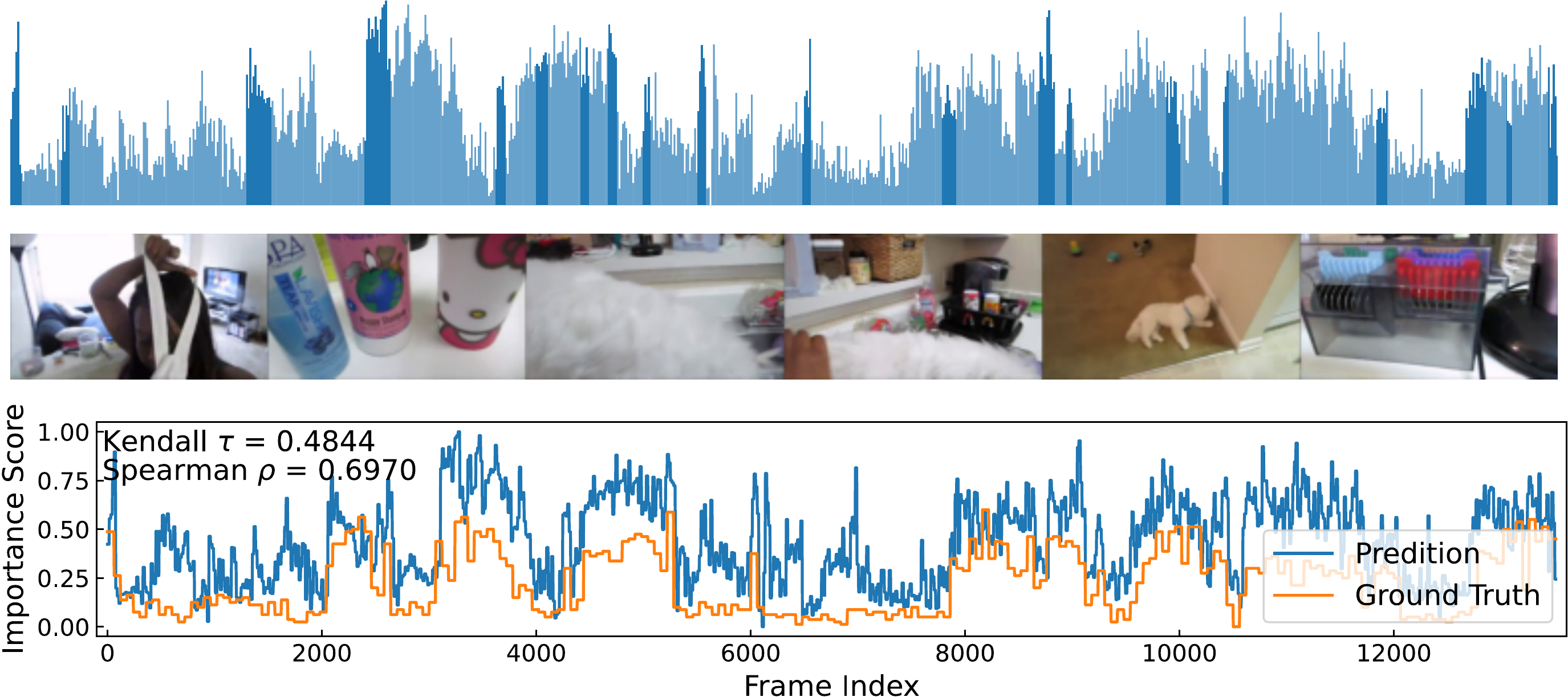}
  \caption{Video name: Bhxk-O1Y7Ho.mp4}
  \label{fig:viz2}
\end{subfigure}

\caption{Visualization of example summaries generated by SUM-DCA. \revised{The height of the light} blue bars indicates ground-truth scores, and dark blue bars denote generated summaries. The blue curves are prediction scores, and the orange curves are the ground truth scores.}
\label{fig:visualize}
\end{figure*}

\subsection{Qualitative Study}
To further intuitively evaluate our model, we visualize some video summaries and the probability curves predicted by SUM-DCA in Figure \ref{fig:visualize}. As we can observe in the figure, the predicted curves are consistent with the human-annotated score curves. In addition, the rank statistics $\tau$ and $\rho$ indicates our model can perform as well as humans. In Figure \ref{fig:viz1}, it demonstrates news about traffic accidents, the selected key shots perfectly summarize the whole news without any redundancy. In Figure \ref{fig:viz2}, the topic of the video is about daily life. Our model selects the diverse shots from the video showing that the effectiveness of diversified contextual attention scheme.

\revised{
\subsection{Discussion}
Our method models contextual information along the video sequence and the diversity among video frames. However, the pairwise similarity measurement is still hand-crafted and cannot be learned during the model optimization. Besides, the partition of video shots still needs to be improved since the quality of the video shots is also important for video summary performance. In future work, we aim to improve the model by incorporating the audio information since we only utilize the visual feature of the video.
}

\section{Conclusion}
This paper has proposed a novel video summarization model named SUM-DCA with diversified contextual attention scheme, which exploits not only global diversity but also local contextual information among video frames. To explore the global diversity, $L_2$ similarity measurement is adopted, which is superior to dot product similarity. Moreover, we utilize the local temporal cue to find the discriminative features through local contextual attention. For proving the effectiveness of the proposed SUM-DCA, we conduct comprehensive experiments as well as the ablation studies on two publicly available data sets. Empirical results have verified that both SUM-DCA and SUM-DCA$_{unsup}$ perform better than other state-of-the-art methods.

\bibliographystyle{unsrt}
\bibliography{egbib}

\begin{thebibliography}{10}

\bibitem{zhang2016video}
Ke~Zhang, Wei-Lun Chao, Fei Sha, and Kristen Grauman.
\newblock Video summarization with long short-term memory.
\newblock In {\em European Conference on Computer Vision}, pages 766--782.
  Springer, 2016.

\bibitem{vaswani2017attention}
Ashish Vaswani, Noam Shazeer, Niki Parmar, Jakob Uszkoreit, Llion Jones,
  Aidan~N Gomez, {\L}ukasz Kaiser, and Illia Polosukhin.
\newblock Attention is all you need.
\newblock In {\em Advances in neural information processing systems}, pages
  5998--6008, 2017.

\bibitem{zhou2018deep}
Kaiyang Zhou, Yu~Qiao, and Tao Xiang.
\newblock Deep reinforcement learning for unsupervised video summarization with
  diversity-representativeness reward.
\newblock In {\em Proceedings of the AAAI Conference on Artificial
  Intelligence}, volume~32, 2018.

\bibitem{jung2019discriminative}
Yunjae Jung, Donghyeon Cho, Dahun Kim, Sanghyun Woo, and In~So Kweon.
\newblock Discriminative feature learning for unsupervised video summarization.
\newblock In {\em Proceedings of the AAAI Conference on Artificial
  Intelligence}, volume~33, pages 8537--8544, 2019.

\bibitem{yuan2019cycle}
Li~Yuan, Francis~EH Tay, Ping Li, Li~Zhou, and Jiashi Feng.
\newblock Cycle-sum: cycle-consistent adversarial lstm networks for
  unsupervised video summarization.
\newblock In {\em Proceedings of the AAAI Conference on Artificial
  Intelligence}, volume~33, pages 9143--9150, 2019.

\bibitem{li2021exploring}
Ping Li, Qinghao Ye, Luming Zhang, Li~Yuan, Xianghua Xu, and Ling Shao.
\newblock Exploring global diverse attention via pairwise temporal relation for
  video summarization.
\newblock {\em Pattern Recognition}, 111:107677, 2021.

\bibitem{zhu2022learning}
Wencheng Zhu, Jiwen Lu, Yucheng Han, and Jie Zhou.
\newblock Learning multiscale hierarchical attention for video summarization.
\newblock {\em Pattern Recognition}, 122:108312, 2022.

\bibitem{zhao2022hierarchical}
Bin Zhao, Maoguo Gong, and Xuelong Li.
\newblock Hierarchical multimodal transformer to summarize videos.
\newblock {\em Neurocomputing}, 468:360--369, 2022.

\bibitem{zhao2018hsa}
Bin Zhao, Xuelong Li, and Xiaoqiang Lu.
\newblock Hsa-rnn: Hierarchical structure-adaptive rnn for video summarization.
\newblock In {\em Proceedings of the IEEE conference on computer vision and
  pattern recognition}, pages 7405--7414, 2018.

\bibitem{mahasseni2017unsupervised}
Behrooz Mahasseni, Michael Lam, and Sinisa Todorovic.
\newblock Unsupervised video summarization with adversarial lstm networks.
\newblock In {\em Proceedings of the IEEE conference on Computer Vision and
  Pattern Recognition}, pages 202--211, 2017.

\bibitem{giles1994dynamic}
C~Lee Giles, Gary~M Kuhn, and Ronald~J Williams.
\newblock Dynamic recurrent neural networks: Theory and applications.
\newblock {\em IEEE Transactions on Neural Networks}, 5(2):153--156, 1994.

\bibitem{hochreiter1997long}
Sepp Hochreiter and J{\"u}rgen Schmidhuber.
\newblock Long short-term memory.
\newblock {\em Neural computation}, 9(8):1735--1780, 1997.

\bibitem{venugopalan2015sequence}
Subhashini Venugopalan, Marcus Rohrbach, Jeffrey Donahue, Raymond Mooney,
  Trevor Darrell, and Kate Saenko.
\newblock Sequence to sequence-video to text.
\newblock In {\em Proceedings of the IEEE international conference on computer
  vision}, pages 4534--4542, 2015.

\bibitem{ji2019video}
Zhong Ji, Kailin Xiong, Yanwei Pang, and Xuelong Li.
\newblock Video summarization with attention-based encoder--decoder networks.
\newblock {\em IEEE Transactions on Circuits and Systems for Video Technology},
  30(6):1709--1717, 2019.

\bibitem{fajtl2018summarizing}
Jiri Fajtl, Hajar~Sadeghi Sokeh, Vasileios Argyriou, Dorothy Monekosso, and
  Paolo Remagnino.
\newblock Summarizing videos with attention.
\newblock In {\em Asian Conference on Computer Vision}, pages 39--54. Springer,
  2018.

\bibitem{zhao2014quasi}
Bin Zhao and Eric~P Xing.
\newblock Quasi real-time summarization for consumer videos.
\newblock In {\em Proceedings of the IEEE conference on computer vision and
  pattern recognition}, pages 2513--2520, 2014.

\bibitem{ngo2003automatic}
Chong-Wah Ngo, Yu-Fei Ma, and Hong-Jiang Zhang.
\newblock Automatic video summarization by graph modeling.
\newblock In {\em Proceedings Ninth IEEE International Conference on Computer
  Vision}, pages 104--109. IEEE, 2003.

\bibitem{liu2002optimization}
Tiecheng Liu and John~R Kender.
\newblock Optimization algorithms for the selection of key frame sequences of
  variable length.
\newblock In {\em European conference on computer vision}, pages 403--417.
  Springer, 2002.

\bibitem{de2011vsumm}
Sandra Eliza~Fontes De~Avila, Ana Paula~Brandao Lopes, Antonio da~Luz~Jr, and
  Arnaldo de~Albuquerque~Ara{\'u}jo.
\newblock Vsumm: A mechanism designed to produce static video summaries and a
  novel evaluation method.
\newblock {\em Pattern Recognition Letters}, 32(1):56--68, 2011.

\bibitem{mei20142}
Shaohui Mei, Genliang Guan, Zhiyong Wang, Mingyi He, Xian-Sheng Hua, and
  David~Dagan Feng.
\newblock L 2, 0 constrained sparse dictionary selection for video
  summarization.
\newblock In {\em 2014 IEEE international conference on multimedia and expo
  (ICME)}, pages 1--6. IEEE, 2014.

\bibitem{rochan2019video}
Mrigank Rochan and Yang Wang.
\newblock Video summarization by learning from unpaired data.
\newblock In {\em Proceedings of the IEEE/CVF Conference on Computer Vision and
  Pattern Recognition}, pages 7902--7911, 2019.

\bibitem{rochan2018video}
Mrigank Rochan, Linwei Ye, and Yang Wang.
\newblock Video summarization using fully convolutional sequence networks.
\newblock In {\em Proceedings of the European Conference on Computer Vision
  (ECCV)}, pages 347--363, 2018.

\bibitem{park2020sumgraph}
Jungin Park, Jiyoung Lee, Ig-Jae Kim, and Kwanghoon Sohn.
\newblock Sumgraph: Video summarization via recursive graph modeling.
\newblock In {\em Computer Vision--ECCV 2020: 16th European Conference,
  Glasgow, UK, August 23--28, 2020, Proceedings, Part XXV 16}, pages 647--663.
  Springer, 2020.

\bibitem{gygli2016video2gif}
Michael Gygli, Yale Song, and Liangliang Cao.
\newblock Video2gif: Automatic generation of animated gifs from video.
\newblock In {\em Proceedings of the IEEE conference on computer vision and
  pattern recognition}, pages 1001--1009, 2016.

\bibitem{hong2020mini}
Fa-Ting Hong, Xuanteng Huang, Wei-Hong Li, and Wei-Shi Zheng.
\newblock Mini-net: Multiple instance ranking network for video highlight
  detection.
\newblock In {\em European Conference on Computer Vision}, pages 345--360.
  Springer, 2020.

\bibitem{ye2021temporal}
Qinghao Ye, Xiyue Shen, Yuan Gao, Zirui Wang, Qi~Bi, Ping Li, and Guang Yang.
\newblock Temporal cue guided video highlight detection with low-rank
  audio-visual fusion.
\newblock In {\em Proceedings of the IEEE/CVF International Conference on
  Computer Vision}, pages 7950--7959, 2021.

\bibitem{xiong2019less}
Bo~Xiong, Yannis Kalantidis, Deepti Ghadiyaram, and Kristen Grauman.
\newblock Less is more: Learning highlight detection from video duration.
\newblock In {\em Proceedings of the IEEE/CVF Conference on Computer Vision and
  Pattern Recognition}, pages 1258--1267, 2019.

\bibitem{badamdorj2021joint}
Taivanbat Badamdorj, Mrigank Rochan, Yang Wang, and Li~Cheng.
\newblock Joint visual and audio learning for video highlight detection.
\newblock In {\em Proceedings of the IEEE/CVF International Conference on
  Computer Vision}, pages 8127--8137, 2021.

\bibitem{szegedy2015going}
Christian Szegedy, Wei Liu, Yangqing Jia, Pierre Sermanet, Scott Reed, Dragomir
  Anguelov, Dumitru Erhan, Vincent Vanhoucke, and Andrew Rabinovich.
\newblock Going deeper with convolutions.
\newblock In {\em Proceedings of the IEEE conference on computer vision and
  pattern recognition}, pages 1--9, 2015.

\bibitem{nair2010rectified}
Vinod Nair and Geoffrey~E Hinton.
\newblock Rectified linear units improve restricted boltzmann machines.
\newblock In {\em Icml}, 2010.

\bibitem{potapov2014category}
Danila Potapov, Matthijs Douze, Zaid Harchaoui, and Cordelia Schmid.
\newblock Category-specific video summarization.
\newblock In {\em European conference on computer vision}, pages 540--555.
  Springer, 2014.

\bibitem{balas1980algorithm}
Egon Balas and Eitan Zemel.
\newblock An algorithm for large zero-one knapsack problems.
\newblock {\em operations Research}, 28(5):1130--1154, 1980.

\bibitem{gygli2014creating}
Michael Gygli, Helmut Grabner, Hayko Riemenschneider, and Luc Van~Gool.
\newblock Creating summaries from user videos.
\newblock In {\em European conference on computer vision}, pages 505--520.
  Springer, 2014.

\bibitem{song2015tvsum}
Yale Song, Jordi Vallmitjana, Amanda Stent, and Alejandro Jaimes.
\newblock Tvsum: Summarizing web videos using titles.
\newblock In {\em Proceedings of the IEEE conference on computer vision and
  pattern recognition}, pages 5179--5187, 2015.

\bibitem{otani2019rethinking}
Mayu Otani, Yuta Nakashima, Esa Rahtu, and Janne Heikkila.
\newblock Rethinking the evaluation of video summaries.
\newblock In {\em Proceedings of the IEEE/CVF Conference on Computer Vision and
  Pattern Recognition}, pages 7596--7604, 2019.

\bibitem{russakovsky2015imagenet}
Olga Russakovsky, Jia Deng, Hao Su, Jonathan Krause, Sanjeev Satheesh, Sean Ma,
  Zhiheng Huang, Andrej Karpathy, Aditya Khosla, Michael Bernstein, et~al.
\newblock Imagenet large scale visual recognition challenge.
\newblock {\em International journal of computer vision}, 115(3):211--252,
  2015.

\bibitem{kingma2014adam}
Diederik~P Kingma and Jimmy Ba.
\newblock Adam: A method for stochastic optimization.
\newblock {\em arXiv preprint arXiv:1412.6980}, 2014.

\bibitem{NEURIPS2019_9015}
Adam Paszke, Sam Gross, Francisco Massa, Adam Lerer, James Bradbury, Gregory
  Chanan, Trevor Killeen, Zeming Lin, Natalia Gimelshein, Luca Antiga, Alban
  Desmaison, Andreas Kopf, Edward Yang, Zachary DeVito, Martin Raison, Alykhan
  Tejani, Sasank Chilamkurthy, Benoit Steiner, Lu~Fang, Junjie Bai, and Soumith
  Chintala.
\newblock Pytorch: An imperative style, high-performance deep learning library.
\newblock In H.~Wallach, H.~Larochelle, A.~Beygelzimer, F.~d\textquotesingle
  Alch\'{e}-Buc, E.~Fox, and R.~Garnett, editors, {\em Advances in Neural
  Information Processing Systems 32}, pages 8024--8035. Curran Associates,
  Inc., 2019.

\end{thebibliography}

\EOD

\end{document}